\documentclass{elektr}
\usepackage{hyperref}
\hypersetup{
colorlinks=true,
urlcolor=blue,
citecolor=blue}
\usepackage[all]{xy,xypic}
\usepackage{amsfonts,amssymb,amsmath,amsgen,amsopn,amsbsy,theorem,graphicx,epsfig}
\usepackage{eufrak,amscd,bezier,latexsym,mathrsfs,eurosym,enumerate}
\usepackage[utf8]{inputenc}
\usepackage[english]{babel}
\usepackage{cleveref,multirow}
\usepackage[dvipsnames]{xcolor}
\usepackage[pagewise]{lineno}

\usepackage{fancyhdr}
\usepackage{tabularx}
\usepackage{booktabs}
\usepackage{graphicx}
\usepackage{subcaption}

\yil{}
\vol{}
\fpage{}
\lpage{}
\doi{}

\title{Electromagnetic Simulations of Antennas on GPUs for Machine Learning Applications}

\author[Murat Temiz, Vemund Bakken]{
\textbf{Murat Temiz$^{1,2}$ Vemund Bakken$^{3}$
\thanks{Murat Temiz}}\\
$^{1}$Department of Electrical and Electronics Engineering, \\ Middle East Technical University, Ankara, Turkey
\\
$^{2}$Department of Electronic and Electrical Engineering, \\ University College London, London, United Kingdom\\
$^{3}$ONiO AS, Oslo, Norway

\\ [1.8em]
}

\def\E{\ifmmode{\mathbb E}\else{$\mathbb E$}\fi} 
\def\N{\ifmmode{\mathbb N}\else{$\mathbb N$}\fi} 
\def\R{\ifmmode{\mathbb R}\else{$\mathbb R$}\fi} 
\def\Q{\ifmmode{\mathbb Q}\else{$\mathbb Q$}\fi} 
\def\C{\ifmmode{\mathbb C}\else{$\mathbb C$}\fi} 
\def\H{\ifmmode{\mathbb H}\else{$\mathbb H$}\fi} 
\def\Z{\ifmmode{\mathbb Z}\else{$\mathbb Z$}\fi} 
\def\P{\ifmmode{\mathbb P}\else{$\mathbb P$}\fi} 
\def\T{\ifmmode{\mathbb T}\else{$\mathbb T$}\fi} 
\def\SS{\ifmmode{\mathbb S}\else{$\mathbb S$}\fi} 
\def\DD{\ifmmode{\mathbb D}\else{$\mathbb D$}\fi} 

\newcommand{\bse}{\begin{subequations}}
\newcommand{\ese}{\end{subequations}}
\newcommand{\ben}{\begin{enumerate}}
\newcommand{\een}{\end{enumerate}}
\newcommand{\bens}{\begin{enumerate*}}
\newcommand{\eens}{\end{enumerate*}}
\newcommand{\be}{\begin{equation}}
\newcommand{\ee}{\end{equation}}
\newcommand{\bea}{\begin{eqnarray}}
\newcommand{\eea}{\end{eqnarray}}
\newcommand{\baa}{\begin{eqnarray*}}
\newcommand{\eaa}{\end{eqnarray*}}
\newcommand{\bc}{\begin{center}}
\newcommand{\ec}{\end{center}}

\theoremstyle{corollary}

\theoremstyle{lemma}

\theoremstyle{proposition}

\theoremstyle{axiom}

\theoremstyle{conjecture}

\theoremstyle{example}

\theoremstyle{definition}

\theoremstyle{remark}

\begin{document}

\maketitle

\abstract{This study proposes an antenna simulation framework powered by graphics processing units (GPUs) based on an open-source electromagnetic (EM) simulation software (gprMax) for machine learning applications of antenna design and optimization. Furthermore, it compares the simulation results with those obtained through commercial EM software. The proposed software framework for machine learning and surrogate model applications will produce antenna data sets consisting of a large number of antenna simulation results using GPUs. Although machine learning methods can attain the optimum solutions for many problems, they are known to be data-hungry and require a great deal of samples for the training stage of the algorithms. However, producing a sufficient number of training samples in EM applications within a limited time is challenging due to the high computational complexity of EM simulations. Therefore, GPUs are utilized in this study to simulate a large number of antennas with predefined or random antenna shape parameters to produce data sets. Moreover, this study also compares various machine learning and deep learning models in terms of antenna parameter estimation performance. This study demonstrates that an entry-level GPU substantially outperforms a high-end CPU in terms of computational performance, while a high-end gaming GPU can achieve around 18 times more computational performance compared to a high-end CPU. Moreover, it is shown that the open-source EM simulation software can deliver similar results to those obtained via commercial software in the simulation of microstrip antennas when the spatial resolution of the simulations is sufficiently fine.}


\maketitle


\section{Introduction}\label{sec1}

EM simulation software and libraries are paramount for antenna and EM research and engineering since the fabrication and measurements of antenna prototypes are relatively expensive and time-consuming. Accordingly, extensive simulations are generally performed to optimize EM structures and antennas before fabricating them. Due to their high reliability and accuracy, commercial EM software packages are widely used in academia and industry to model, design, and optimize antennas. However, commercial EM simulation software packages are generally expensive and are not suitable for producing data sets consisting of a large number of simulation results, which are required for machine learning applications, since these software packages are closed-source and do not generally provide a scripting interface that can be used to automize the simulation process and data processing. On the other hand, various open-source EM simulation software packages and libraries have been developed and utilized in research due to their flexibility, open-source structure, and editable libraries, and hence, further development opportunities. In \cite{ref1, ref2}, researchers investigated the accuracy and computational performance of various open-source antenna simulation software packages and libraries, and they presented a comparison of them with commercial EM simulation software packages. They considered nec2c \cite{ref3}, gprMax \cite{ref4}, openEMS \cite{ref5} in comparison with analytical solutions of antennas and two commercial software suites (e.g., CST Studio, HFSS). These studies demonstrated that open-source software packages can achieve high accuracy and performance in the simulation of well-known antenna structures such as dipole, patch, and loop antennas.

In this study, we have utilized an open-source EM simulation package, namely gprMax, which has been mainly developed for ground penetrating radar (GPR) simulations. However, it may also be used as a general EM simulation tool, such as for simulating antenna models or EM waves \cite{ref4}. This open-source package utilizes the finite-difference time-domain (FDTD) method to simulate EM waves and their interactions with the materials. Developers of gprMax also modelled commercial GPR antennas and investigated their accuracy and performance in gprMax \cite{ref6,ref7,ref8}. These commercial GPR antenna modules consist of two bowtie antennas placed in a case, and they operate at 1.5 or 1.2 GHz central frequencies. Their investigation demonstrated that gprMax achieves a sufficient accuracy at 1.2 GHz and 1.5 GHz frequencies in comparison with CST simulations and measurement results \cite{ref7}. Moreover, in another study \cite{ref9}, researchers considered three different types of antennas and demonstrated that gprMax can be effectively used for simulation of the bowtie, Vivaldi, and dipole antennas. They compared the simulation results with measurements performed in the lab. When the complexity of the antenna design increases, the gap between the simulations and measurements becomes more significant, as observed in the case of the Vivaldi antenna in \cite{ref9}. However, the simulation results are still promising, and they can be further improved by tuning the FDTD parameters, as we have shown in this study.

Designing an optimum antenna to meet the requirements of specific applications, such as Internet of Things (IoT) nodes, smartphones, and ultra-wideband (UWB) applications within a limited space of devices might be challenging. Moreover, it may need to perform a large number of simulations. For instance, the FDTD method requires a large amount of memory and computational resources to simulate large antenna structures. Consequently, the accuracy of the simulation software and its computing performance are crucial to producing a cutting-edge and optimum design while reducing the design, optimization, and simulation time. Therefore, various techniques, such as machine learning or surrogate models, are also utilized for antenna design and optimization.

{Machine learning and deep learning-based techniques have been recently developed for antenna design and optimization since these techniques can significantly reduce the computational complexity and provide near-optimum results \cite{ref10, ref11}. Recent studies have focused on applications of machine learning and deep learning techniques for solving EM problems, designing and optimizing antennas \cite{ref12,ref13}.} Shi et al. proposed a framework based on ensemble learning that utilizes random forest, linear regression, support vector machine, and gradient boosting techniques for smart antenna classification and antenna geometric parameter prediction \cite{ref14}. Another study proposes a technique based on particle swarm optimization and neural networks for the design and optimization of antennas \cite{ref15}. Haque et al. utilized random forest and regression methods in addition to EM simulations to design a 28 GHZ antenna for 5G wireless networks \cite{ref16}. Multi-port antenna design for MIMO systems via neural networks and random forest is proposed in \cite{ref17}. In addition to EM simulations, surrogate models are also considered with machine learning techniques to swiftly optimize antennas \cite{ref18}.  Another study also utilized surrogate models with a convolutional neural network (CNN) and Gaussian process regression (GPR) to optimize and design antennas \cite{ref19}.

Moreover, machine learning techniques-based surrogate models were also considered for eliminating the need for repeating computationally expensive EM simulations  \cite{ref20,ref21}. Surrogate models were established based on the existing simulation results of antennas, and they are utilized to design and optimize new antennas using different optimization algorithms. However, establishing surrogate models and training machine learning algorithms requires data sets consisting of a large number of simulated antennas (e.g., {$>1000$}) with their simulation results. It was shown that the general-purpose computing on GPUs (GPGPU) can substantially accelerate the FDTD method by concurrently running the simulation on plenty of compute unified device architecture (CUDA) cores~\cite{ref21a}. GPUs consist of a large number of small parallel processing cores, while CPUs consist of a few powerful multipurpose cores \cite{ref21b}. The FDTD method can be concurrently executed on a large number of computational cores of GPUs because the electric and magnetic fields can be independently updated at each point of the computational grid in the FDTD method during simulations \cite{ref22}. Moreover, multiple GPU cores can simultaneously access the local RAM of the graphics card at ultra-high data rates, maximizing computation performance by avoiding bottlenecks between the RAM and CUDA cores. Accordingly, it is important to utilize GPUs in FDTD simulations, which can tremendously reduce the simulation time.

\begin{table}[h]
\centering

\caption{Summary of Literature on Antenna Simulation, Optimization, and Machine Learning.}\label{tab:literature}
\renewcommand{\arraystretch}{1.3}
\renewcommand{\arraystretch}{1.3}
\begin{tabular}{|p{0.8cm}|p{1.1cm}|p{2.9cm}|p{4.2cm}|p{5.8cm}|}
\hline
\textbf{Year} & \textbf{Study} & \textbf{Method(s)} & \textbf{Application / Focus} & \textbf{Key Findings} \\
\hline
2014 & \cite{ref20} & SM + ML & EM simulation acceleration & SMs avoid repeated FDTD runs. \\
\hline
2015 & \cite{ref6} & gprMax (FDTD), CST & GPR bowtie antennas & gprMax models vs CST simulations. \\
\hline
2016 & \cite{ref4, ref7} & gprMax (FDTD), Measurements & GPR Antennas at 1.2/1.5 GHz & gprMax models vs CST and lab data. \\
\hline
2017 & \cite{ref9} & gprMax + Measurements & Bowtie, Vivaldi, Dipole antennas & Accuracy drops for complex antennas (e.g., Vivaldi). \\
\hline
2020 & \cite{ref12, ref13} & ML & EM problem solving & Effective ML-based optimization. \\
\hline
2020 & \cite{ref10} & ML & Antenna design & Reduced complexity, near-optimal results. \\
\hline
2020 & \cite{ref21} & SM + ML & Antenna optimization & SMs reduce EM simulation time. \\
\hline
2022 & \cite{ref14} & Ensemble Learning (RF, SVM, GB) &  Antenna parameter prediction & Accurate prediction of geometry and type. \\
\hline
2023 & \cite{ref11} & DL & Antenna optimization & DL improves simulation speed and accuracy. \\
\hline
2024 & \cite{ref15} & PSO + NN & Antenna optimization & Efficient hybrid ML design framework. \\
\hline
2024 & \cite{ref17} & NN, RF & MIMO multi-port antennas & ML enables efficient MIMO design. \\
\hline
2024 & \cite{ref18} & SM + ML & Antenna optimization & SMs speed up ML-based design. \\
\hline
2024 & \cite{ref19} & CNN + Regression + SM & Antenna design & ML and SM combination improves design speed and accuracy. \\
\hline
2025 & \cite{ref16} & RF, Regression + EM Simulation & 5G antenna (28 GHz) & ML-assisted mmWave antenna design. \\
\hline
\end{tabular}
\bigskip
\begin{minipage}{\textwidth}
\footnotesize
\textbf{Acronyms:} ML – Machine Learning; DL – Deep Learning; FDTD – Finite-Difference Time-Domain; GPR – Ground Penetrating Radar; RF – Random Forest; PSO – Particle Swarm Optimization; CNN – Convolutional Neural Network; Regression – Statistical Regression; EM – Electromagnetic; NN – Neural Networks; SVM – Support Vector Machine; LR – Linear Regression; GB – Gradient Boosting; SM – Surrogate Model.
\end{minipage}
\end{table}

Table~\ref{tab:literature} provides a summary of the literature on antenna simulation, optimization, and machine learning applications. Previous studies considered less complex and well-known antenna structures, such as dipole or Vivaldi antennas, for the simulations using gprMax \cite{ref1,ref2,ref9}. The application of machine learning and surrogate models requires a large amount of training data; obtaining this training data is computationally challenging using commercial EM simulation software packages. Moreover, modern GPUs substantially outperform CPUs for FDTD simulations in terms of speed. Motivated by these reasons, this study aims to examine the accuracy of an open-source EM simulation library, i.e., gprMax, on the simulation of various antenna structures to generate large-scale data sets for machine learning and surrogate model applications. Moreover, we also evaluate the computational performance of GPUs and CPUs to provide a comparison between them in terms of EM simulations. Therefore, the contributions of this study are as follows;

\begin{itemize}
    \item We propose a simulation framework and specifically focus on simulating more sophisticated antenna structures using gprMax on GPUs and compare. This study also evaluates its accuracy against commercial EM simulation software.
    \item  This study also evaluates the computational performance of GPU and CPU for sophisticated antenna simulations using gprMax and provides a comparison of them.
    \item  We also provide the application of various machine learning models and their accuracy for the prediction of antenna geometry parameters using the dataset obtained via the proposed simulation framework.
\end{itemize}

{In the remainder of} the paper, Section II, introduces the system model, including the simulation model utilizing GPUs. At the same time, Section III presents example antenna models that are used to evaluate gprMax and GPU computing. Section IV compares the simulation results and performance of the CPU and GPUs. Finally, the conclusions are drawn in Section V.

\section{System Model}\label{sys_model}
This section explains how antenna simulations are performed in gprMax using GPUs and provides the mathematical model of the simulation framework. gprMax utilizes the FDTD method to simulate electromagnetic waves, and it has been mainly developed for numerical modelling and simulating Ground Penetrating Radar (GPR) applications and scenarios \cite{ref4}. Moreover, the recent versions of gprMax support CUDA-based calculations for GPR simulations \cite{ref22}. It also supports the simulation of antennas via the transmission line feed model provided in the gprMax library. However, this transmission line feed model does not currently support GPU CUDA for antenna simulations. For this reason, we have employed \textit{voltage source} (Vs) and \textit{receiver} (Rx) models of gprMax to excite the antenna and receive the reflected signals, respectively. Since we aim to calculate the S11 of the antenna, Vs and Rx are connected to the same feed location (antenna input port). After that, the signals received from the antenna feed are processed to calculate the S11 (reflection coefficient) of the antenna simulated. It is worth noting that a comparison of common FDTD feed models can be found in \cite{ref23}, where the hard-source and transmission-line feed models were examined and compared.

\subsection{FDTD Simulations}

The FDTD method numerically solves Maxwell's equations in the time domain to simulate EM waves and structures. It works by discretizing both time and space, allowing the simulation of how electromagnetic fields evolve and interact with materials. FDTD is based on the time-dependent form of Maxwell’s curl equations \cite{ref24},
\begin{equation}
\begin{aligned}
\nabla \times \mathbf{E} = -\mu \frac{\partial \mathbf{H}}{\partial t}, \quad\quad\quad
\nabla \times \mathbf{H} = \epsilon \frac{\partial \mathbf{E}}{\partial t},
\end{aligned}
\end{equation}
where $\mathbf{E}$ is the electric field, $\mathbf{H}$ is the magnetic field, $\mu$ is the permeability, and $\epsilon$ is the permittivity of the medium. The computational domain is divided into a grid known as the Yee lattice \cite{ref25}, where the electric and magnetic field components are computed at alternating time steps to improve numerical stability and accuracy.

To implement FDTD, the spatial and temporal derivatives in Maxwell's equations are approximated using central difference formulas. The fields are updated alternately, such that firstly the magnetic field components are updated using the electric fields from the previous time step, then the electric fields are updated using the new magnetic fields. This explicit time-stepping scheme continues until the simulation is completed for the desired time. To ensure numerical stability in FDTD simulations, the time step $\Delta t$ must satisfy the Courant-Friedrichs-Lewy (CFL) condition, which is given by,
\begin{equation}
\Delta t \leq \frac{1}{c \sqrt{\left(\dfrac{1}{\Delta x^2} + \dfrac{1}{\Delta y^2} + \dfrac{1}{\Delta z^2} \right)}},
\end{equation}
where $c = 1/\sqrt{\mu \epsilon}$ is the speed of light in the medium, and $\Delta x$, $\Delta y$, and $\Delta z$ are the spatial steps in each direction. This condition ensures that the simulation remains stable and accurately captures the wave propagation over time.

\subsection{gprMax and Antenna Simulations using CUDA}

Vs and Rx elements are connected at the same location (antenna feed point) such that Vs transmits a single Gaussian pulse, denoted by vector $\mathbf{V}_t$, with a centre frequency of $f_c=2$ GHz that can cover the simulation of the antennas up to 6 GHz. The internal resistance of Vs, i.e., port impedance, is set to 50 $\Omega$, and its polarization is in the direction of the antenna feed line. Therefore, $\mathbf{V}_t$ is expressed as

\begin{equation}
    \mathbf{V}_t = \exp{(-2\pi^2f_c ^2(\mathbf{t}_s-\frac{1}{f_c})^2)},
\end{equation}
where $\mathbf{t}_s$ contains the time samples of the simulation.

The vectors containing measured electric fields at the Rx location are calculated by solving the full FDTD equations and returned as $\mathbf{E}_x$, $\mathbf{E}_y$, and $\mathbf{E}_z$ for x, y, and z polarizations, respectively. Afterward, the received voltage vectors in each polarization can be calculated as
\begin{equation}
    \mathbf{V}_{r,x}=-\mathbf{E}_{x}d_{x}, 
\end{equation}
\begin{equation}
    \mathbf{V}_{r,y}=-\mathbf{E}_{y}d_{y}, 
\end{equation}
\begin{equation}
    \mathbf{V}_{r,z}=-\mathbf{E}_{z}d_{z}, 
\end{equation}
where $\mathbf{V}_{r,x}$, $\mathbf{V}_{r,y}$, $\mathbf{V}_{r,z}$ denote the calculated amplitude vectors at the Rx in x, y, and z polarizations, respectively.  $\mathbf{E}_{x}$, $\mathbf{E}_{y}$, $\mathbf{E}_{z}$ denote the electric fields in each polarization at the Rx, which are calculated by gprMax via the FDTD method. Furthermore, $d_{x}$, $d_{y}$ and $d_{z}$ denote the FDTD cell size in x, y, z directions, respectively. Not that calculated amplitudes also include the direct signals from the Vs in addition to the signals reflected by the antenna. Accordingly, the vectors containing the reflected signals from the antenna in x polarization are calculated as
\begin{equation}
    \mathbf{V}_{a,x} = \mathbf{V}_{r,x} - \frac{1}{2}\mathbf{V}_{t,x},
    \label{reflected_signal}
\end{equation} where $\mathbf{V}_{t,x}$ denotes the  x polarization component of excitation signal $\mathbf{V}_{t}$. Moreover, equation~(\ref{reflected_signal}) is also valid for reflected signals in the y and z polarizations. It should be noted that excitation signals $\mathbf{V}_{t}$ have a single polarization defined in the FDTD model, so that one of the polarizations is equal to $\mathbf{V}_{t}$ while others are 0. For instance, if voltage source Vs is defined in the x polarization then $\mathbf{V}_{t,x}=\mathbf{V}_{t}$, $\mathbf{V}_{t,y}=0$, and $\mathbf{V}_{t,z}=0$.

For s-parameter calculations, calculating only the polarization in which Vs excites the antenna is sufficient. Assuming that Vs excites the antenna in the x polarization, the S11 of the antenna in terms of transmitted and reflected signal amplitudes is then calculated as 

\begin{equation}
    S_{11} = \left|\frac{\mathcal{FFT}\left(\mathbf{V}_{a,x}\right)}{\mathcal{FFT}\left(\mathbf{V}_{t,x}\right)}\right|,
\end{equation}
where  $\mathcal{FFT\left(.\right)}$ defines the Fast Fourier transform (FFT) operator. Eventually, the $S_{11}$ of the antenna in dB can is given by 
\begin{equation}
    S_{11}{[dB]}=20\log_{10}(S_{11}).
\end{equation}
By employing the method given above, antenna simulations can be performed on CUDA-enabled GPUs in gprMax. Moreover, this method can be modified for calculating other s-parameters, e.g., $S_{12}$, if there is more than one antenna port in the simulations. However, this is beyond the scope of this paper, and it will be investigated in a future study.

\subsection{Simulation Framework and Database Model}
The antenna simulation framework has been developed in two parts, both of which are written in Python. The first part is to select the simulation parameters, antenna geometry, process the simulation data, store the final results in the database, and sketch the necessary figures. The second part of the framework, which is the simulation core, simulates the antenna structure utilizing the gprMax and records the raw simulation data in the database. This enables a flexible software architecture for simulating various antennas. Perfectly Matched Layers (PML) are used as the boundary condition. PML is an absorbing boundary that simulates an infinite space, preventing reflections at the edges of the computational domain. The simulation uses manual mesh refinement by selecting different uniform grid sizes ($dx$, $dy$, $dz$) for each model run, allowing evaluation of accuracy versus computational cost, as shown in the results. No adaptive or local mesh refinement is applied; instead, entire simulations are run with finer or coarser grids to investigate their impact on the accuracy and computational complexity.

The hierarchical data format version 5 (HDF5) is used to store a large number of simulation results due to its capabilities of handling large, complex, heterogeneous data sets \cite{ref26}. This database format stores the entire data in a compressed single data file, which makes the database portable and easy to share. The HDF5 database format is widely used in various scientific communities. The H5py package is used in Python to manage the HDF5 database in this study. The parameters of the antenna designs and their corresponding simulation results are stored in a single HDF5 file to enable easy integration of the simulation results with machine learning applications or other data-driven techniques such as surrogate modeling.
\section{Example Antenna Models}\label{AntennaModels}

This section investigates the simulations of three different antennas using gprMax on CPU and GPUs. Moreover, it compares the results to the ones obtained via a commercial antenna simulation software, namely CST Microwave Studio \cite{ref27}. It is worth noting that CST Microwave Studio simulates the antennas using the finite integration technique (FIT) in the time domain. Moreover, CST is highly optimized for antenna simulations to enhance and accelerate the accuracy of simulations; therefore, it is considered a benchmark in this study.

The simulations of simpler antennas via gprMax were previously investigated in \cite{ref1,ref2,ref9}; accordingly, we have opted for more complicated antenna structures, namely, inverted-F antenna, multi-band inverted-F antenna, and multi-band dipole antenna, to evaluate the accuracy and performance of gprMax with more sophisticated antenna structures. 

\subsection{IFA Antenna}
\begin{figure}
\centering{}\includegraphics[width=0.6\columnwidth]{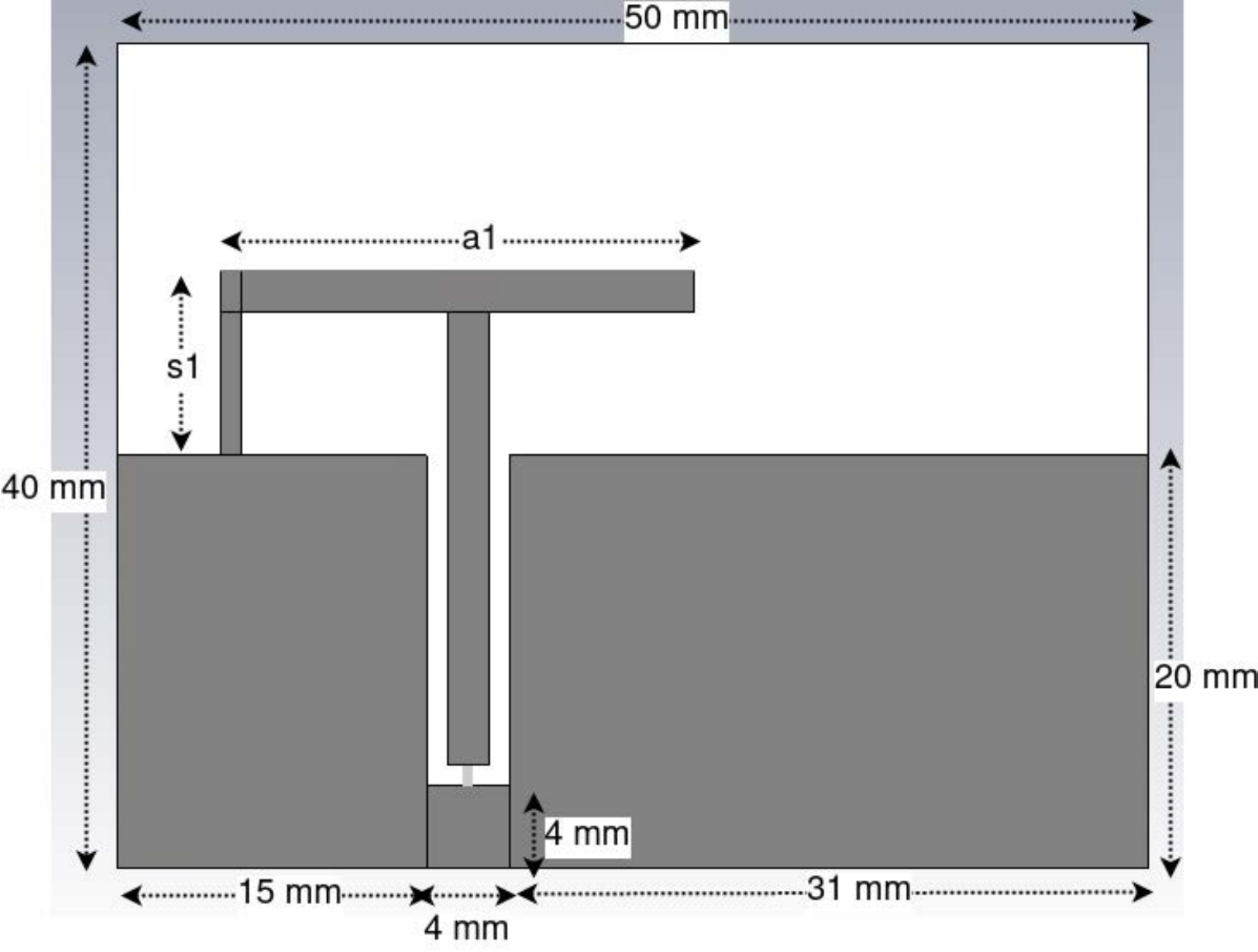}\vspace{-0.05in}\caption{Illustration of the IFA model.  \label{fig:ifa}}
\end{figure}

Inverted-F (IFA) antennas are commonly used in wireless and mobile communication devices such as mobile phones, Internet of Things (IoT) nodes, or portable wireless devices due to their small size, lightness, and robust structure \cite{ref28,ref29}. One end of the IFA antenna is connected to the ground plane while the other end is open, and the feed is located between these two ends. The main advantage of an IFA antenna is its shorter physical length ($<1/4\lambda$) compared to the wavelength ($\lambda$), and easy and cost-effective fabrication on the PCB that can also contain the electronic circuit. A coplanar waveguide excited IFA model has been investigated in this study. Coplanar waveguide excited IFA was shown to provide wider bandwidth, and it is also easier for integration and fabrication compared to microstrip excited IFA since the waveguide, antenna, and ground plane are on only a single PCB layer \cite{ref30}. Fig~\ref{fig:ifa} illustrates the IFA considered in this study with its size parameters. The substrate is a $50\times40$ mm$^2$ FR-4 PCB with a thickness of 1 mm. The width of the coplanar waveguide feed line is 2 mm, and the width of the short line between the antenna and the ground plane is 1 mm. The parameters $a1$ and $s1$, can be changed to produce various operating frequencies and bandwidths. In this investigation, $a1=30$ mm and $s1=12$ mm are chosen. The simulation area of the FDTD model is $80\times80\times40$ mm$^3$. The FDTD cell size is defined by $dx$, $dy$, and $dz$ in the x, y, and z directions.

\subsection{Dual-band IFA Antenna}
\begin{figure}
\centering{}\includegraphics[width=0.6\columnwidth]{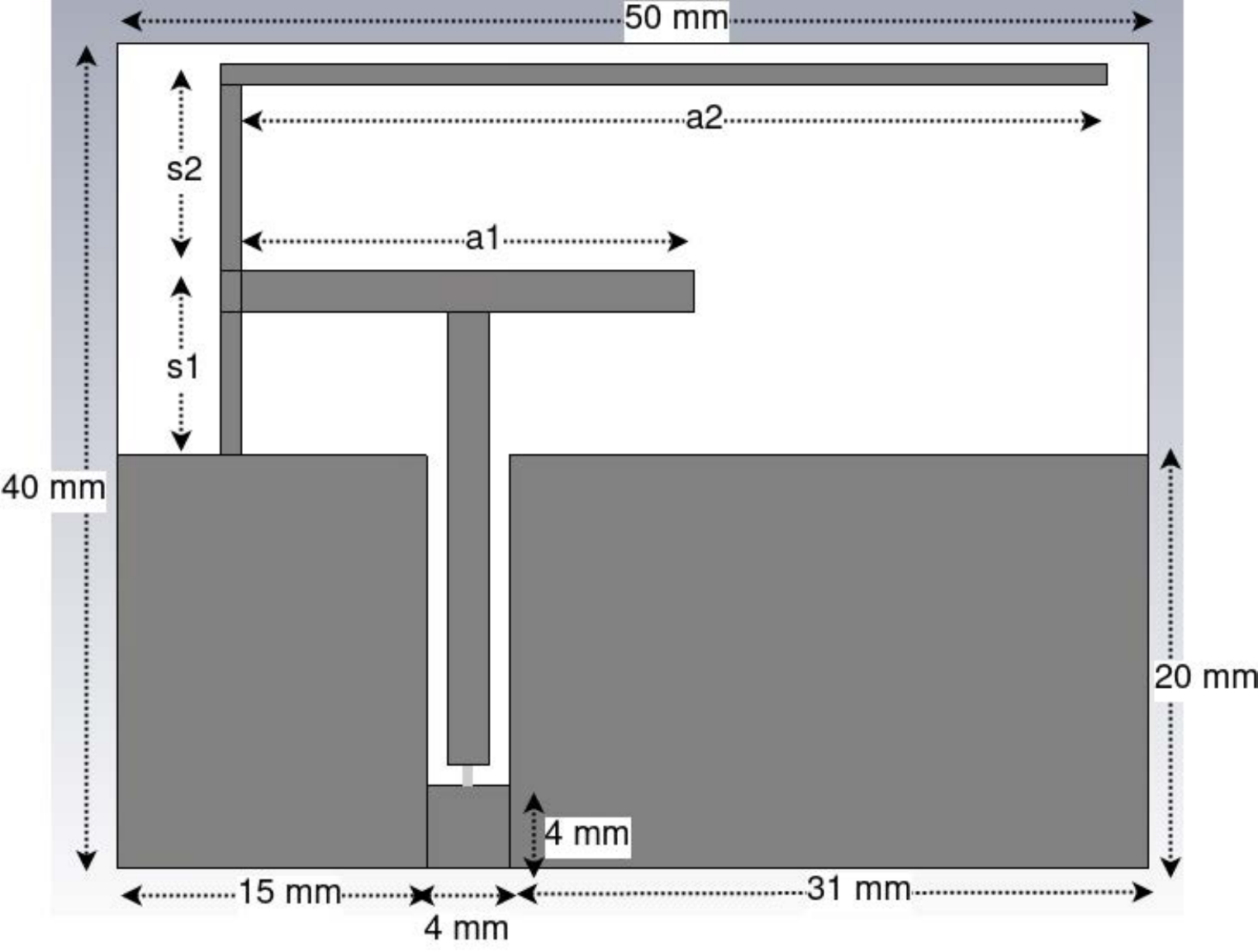}\vspace{-0.05in}\caption{Illustration of the dual-band IFA model.  \label{fig:dual_ifa}}
\end{figure}

This antenna is obtained by adding an extra arm to the IFA, which provides an additional narrow-band resonance frequency in addition to the wideband response of the IFA presented in the previous section. Fig.~\ref{fig:dual_ifa} illustrates the model of this antenna and presents size parameters $a1$, $a2$, $a3$, and $a4$, which can be varied to produce various operation frequencies. In this study, $a1=30$ mm, $a2=46$ mm, $s1=12$ mm and $s2=8$ mm are chosen. The size of the second arm, $a2$, mainly determines the lower narrow-band operating frequency of the antenna, such that having a larger $a2$ decreases the lower narrow-band resonance frequency. However, this also impacts the wideband frequency band as a result of EM interaction between the arms, and this is discussed in the simulation results section. The simulation area of the FDTD model is $80\times80\times40$ mm$^3$, the same as the IFA presented in the previous section. 

\subsection{Multi-band Dipole Antenna}
\begin{figure}
\centering{}\includegraphics[width=0.7\columnwidth]{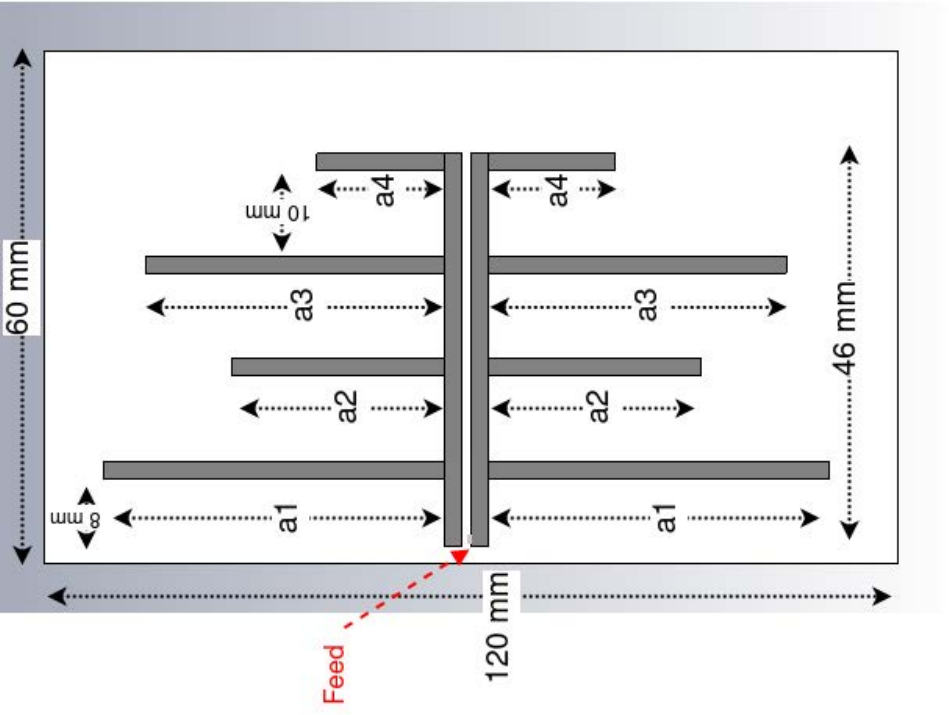}\vspace{-0.05in}\caption{Illustration of the multi-band dipole antenna model.  \label{fig:multiband_dipole}}
\end{figure}

In addition to the IFA models presented in the previous sections, a multi-band dipole antenna structure consisting of 4 dipole pairs is modelled and simulated. Having 4 dipole pairs, this antenna can produce multiple resonance frequencies, and these resonance frequencies are related to the lengths of the dipole arms. Fig.~\ref{fig:multiband_dipole} illustrates the multi-band dipole antenna model, where
parameters $a1$, $a2$, $a3$, and $a4$ can be varied to produce various resonance frequencies. A similar antenna was previously proposed to operate in multiple frequency bands for mobile communication and navigation applications in \cite{ref31}, where it was experimentally shown that this type of antenna can effectively cover multiple frequency bands, including GSM, LTE, WiFi, 4G, and GPS bands with reasonable gain and bandwidth. 

In this study, $a1=40$ mm, $a2=25$ mm, $a3=35$ mm and $a4=15$ mm are chosen. The simulation area of the FDTD model of this antenna is $80\times140\times28$ mm$^3$, which is larger than the simulation area of the previous antennas. Hence, it is expected that the simulation of this antenna will require a larger amount of RAM compared to the IFA antennas.

\section{Simulation Results and  Performance}\label{Numerical}

\begin{figure}
\centering{}\includegraphics[width=0.75\columnwidth]{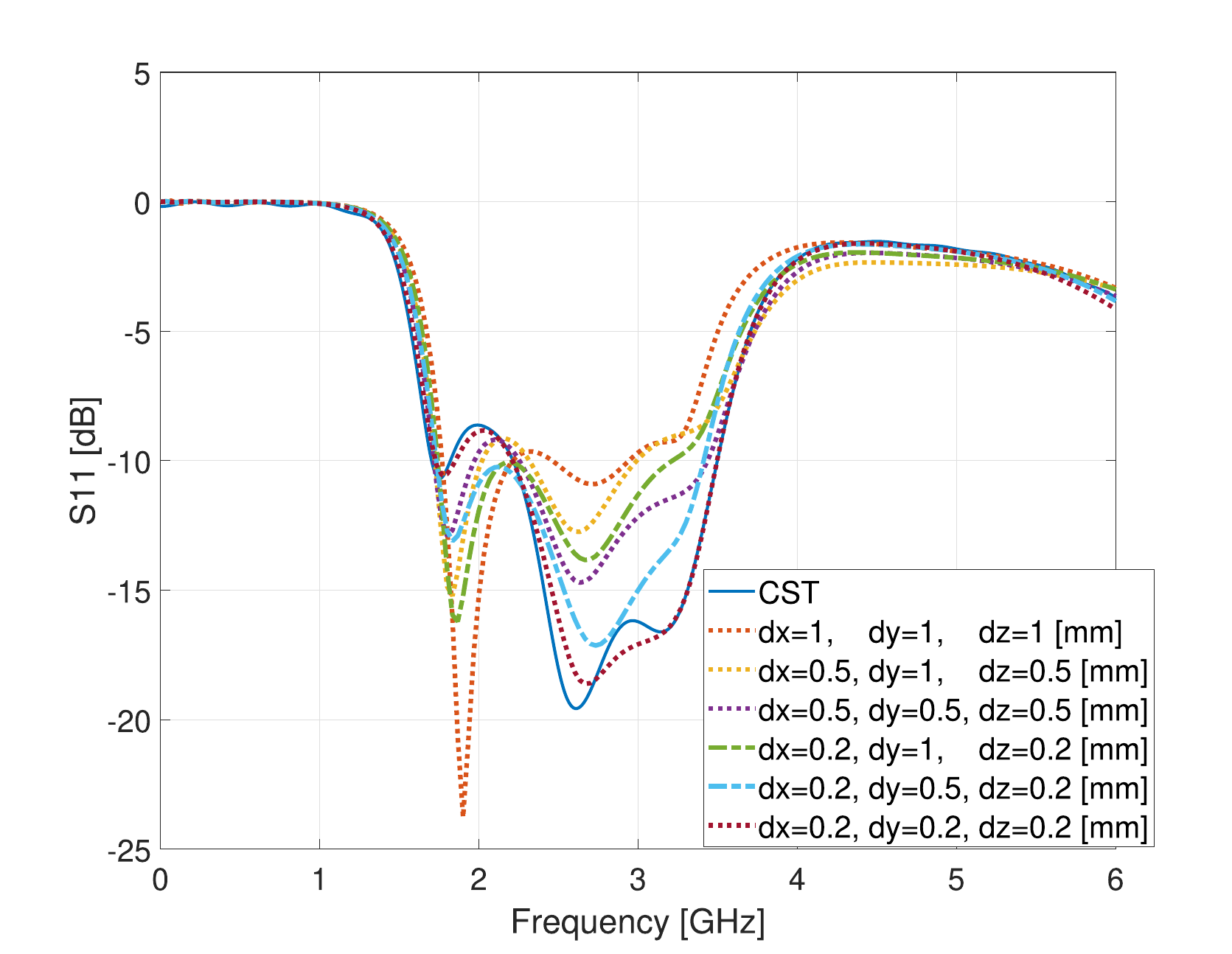}\vspace{-0.05in}\caption{Comparison of IFA S11 simulations via CST and gprMax with various FDTD cell sizes. \label{fig:s11_ifa}}
\end{figure}

\begin{figure}
\centering{}\includegraphics[width=0.75\columnwidth]{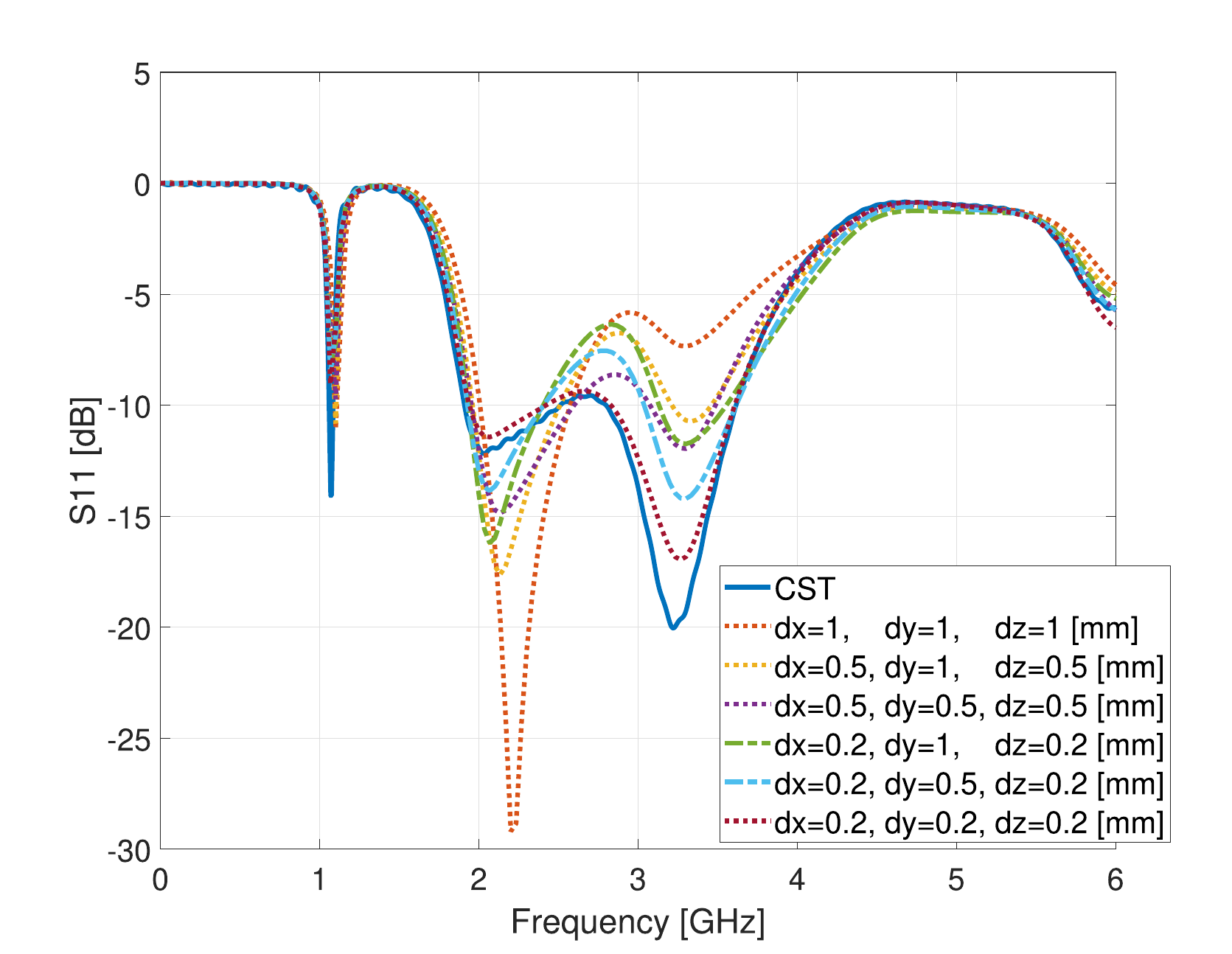}\vspace{-0.05in}\caption{Comparison of dual-band IFA S11 simulations via CST and gprMax with various FDTD cell sizes. \label{fig:s11_dualifa}}
\end{figure}

\begin{figure}
\centering{}\includegraphics[width=0.75\columnwidth]{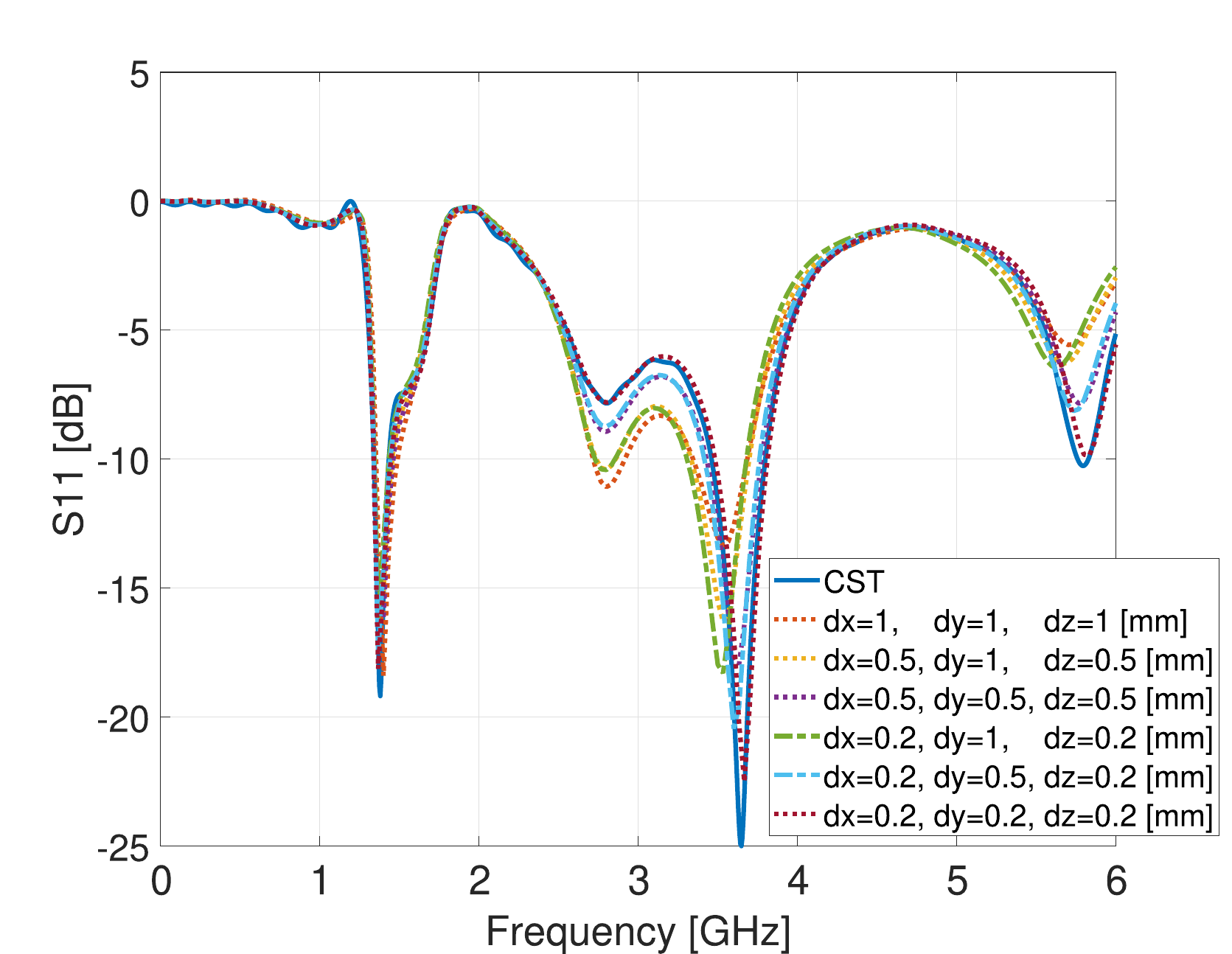}\vspace{-0.05in}\caption{Comparison of multi-band dipole antenna S11 simulations via CST and gprMax with various FDTD cell sizes.\label{fig:s11_multidipole}}
\end{figure}

This section presents the antenna simulation results and discusses them while also comparing the performance of CPUs and GPUs in antenna simulations.

\subsection{Antenna Simulation Results}

\begin{table*}[h]
\caption{Comparison of computing hardware used in antenna simulations.}
\centering
\begin{tabular}{||c|c|c|c|c||} 
 \hline
 Hardware & Processing Unit & RAM & Max Clock & Transistors \\ [0.5ex] 
 \hline\hline
 Ryzen 9 3900X & 12 cores (24 Threads) & 32 GB (System RAM) & 4.6 GHz & 3,8 Billion
 \\ 
 \hline
 Geforce RTX 3070 & 5888 CUDA Cores & 8 GB & 1.725 GHz & 17.4 Billion \\
 \hline
 Quadro P620 & 512 CUDA Cores & 2 GB & 1.35 GHz & 3.3 Billion\\
 \hline
\end{tabular}
\label{table:hardware}
\end{table*}

For each antenna considered, the size of an FDTD cell is varied from $dx = 0.2$ mm, $dy=0.2$ mm, $dz=0.2$ mm to $dx = 1$ mm, $dy=1$ mm, $dz=1$ mm. It is observed that having a finer FDTD cell size makes the gprMax simulation results closer to the ones obtained by CST Studio. {Fig.~\ref{fig:s11_ifa}, Fig.~\ref{fig:s11_dualifa}, and Fig.~\ref{fig:s11_multidipole} compare the S11 of the simulations obtained by the proposed method in comparison with the ones obtained via the benchmark method, i.e., CST Studio. It is seen that choosing a smaller FDTD cell size improves the accuracy of the simulations, as explained below in detail for each antenna.}

Specifically, Fig.~\ref{fig:s11_ifa} presents the S11 results obtained via gprMax and CST for the IFA with $a1=30$ mm and $s1=12$ mm. It presents that when the FDTD cell is $dx=0.2$ mm, $dy=0.2$ mm, and $dz=0.2$ mm, S11 results obtained via gprMax and CST Studio are very close. On the other hand, having a larger FDTD cell size degrades the accuracy of the gprMax simulations, although it is still acceptable with $dx=0.5$ mm, $dy=0.5$ mm, and $dz=0.5$ mm cell size. A similar trend has been observed in the simulation of dual-band IFA antenna, especially in the higher frequencies, as shown in Fig.~\ref{fig:s11_dualifa}. However, for the lower frequency part of the simulation, i.e., < 1 GHz, a good matching between gprMax and CST has been seen even with a coarse FDTD cell size, i.e., $dx=1$ mm, $dy=1$ mm, and $dz=1$ mm, as seen in Fig.~\ref{fig:s11_dualifa}. This is consistent with the FDTD simulation theory since the FDTD size should be determined concerning the wavelength of the maximum simulated frequency, which means that for simulations with lower frequency (<1 GHz), FDTD cell sizes with $dx=1$ mm, $dy=1$ mm, and $dz=1$ mm should be sufficient. By comparing Fig.~\ref{fig:s11_ifa} and Fig.~\ref{fig:s11_dualifa}, it is observed that adding the second resonance arm to the IFA antenna to obtain the dual-band IFA also affects the wideband response of the antenna since there is an interaction between the first and second arms of the dual-band IFA.

The multi-band dipole antenna presented in Fig.~\ref{fig:multiband_dipole} covers a wider frequency band and multiple resonance frequencies up to 6 GHz, as shown in Fig.~\ref{fig:s11_multidipole}. For this antenna, a nearly perfect match between the results obtained via CST and gprMax is noticed when the FDTD cell size of gprMax is $dx=0.2$ mm, $dy=0.2$ mm, and $dz=0.2$ mm. Moreover, even with a coarser FDTD cell size, gprMax still performs reasonably well on these antenna models. However, the discrepancy between the accuracy of simulations with coarse and finer FDTD cell sizes gets more significant as the frequency of the simulations increases. {These results show that the proposed framework works well and its accuracy is comparable to the commercial EM simulation packages, even for sophisticated antenna simulations.}

\begin{figure}[ht]
    \centering
    \begin{subfigure}[b]{0.3\textwidth}
        \centering
        \includegraphics[width=\textwidth]{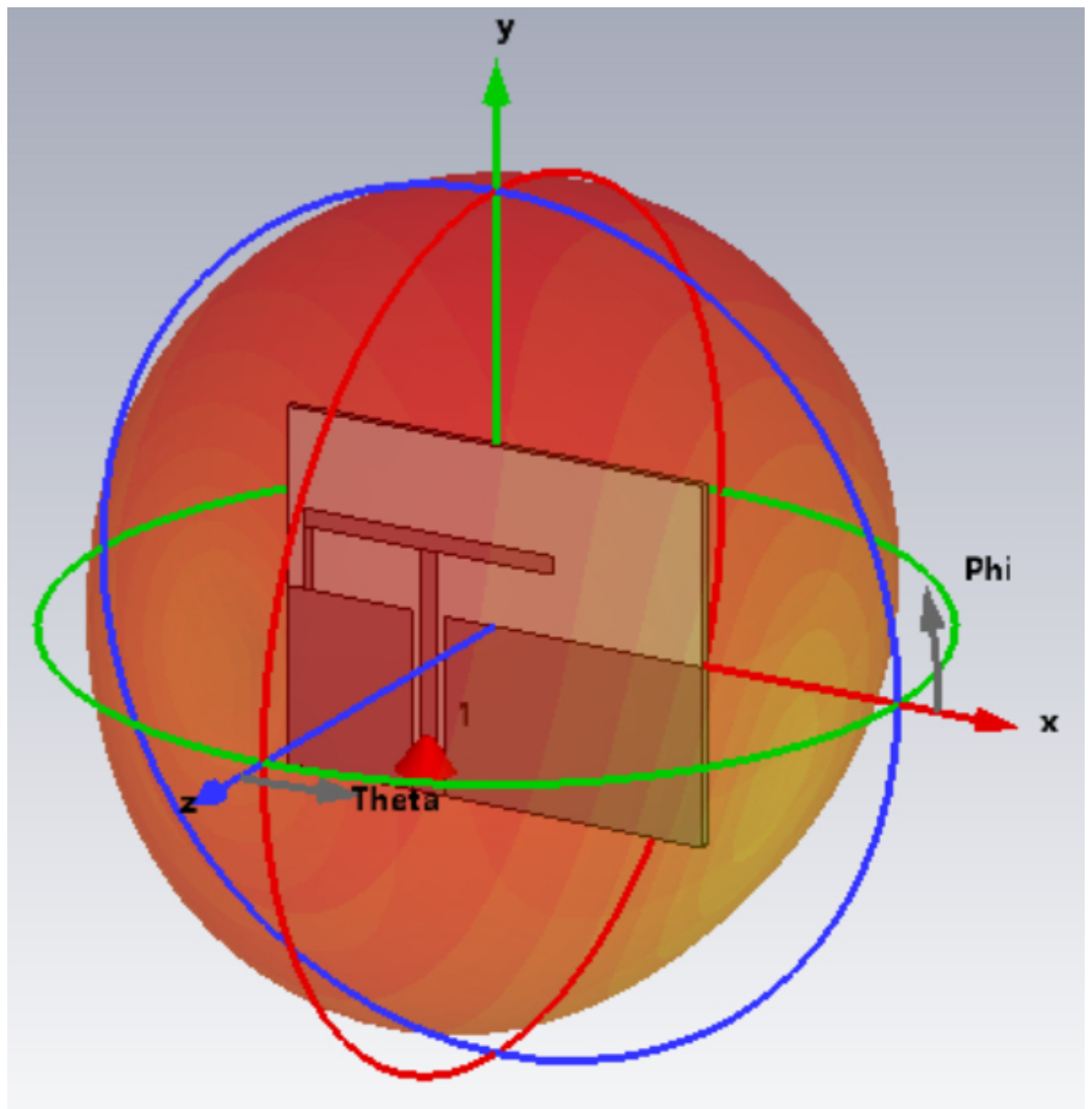}
        \caption{IFA at 2.6 GHz}
        \label{fig:sub1}
    \end{subfigure}
    \hfill
    \begin{subfigure}[b]{0.3\textwidth}
        \centering
        \includegraphics[width=\textwidth]{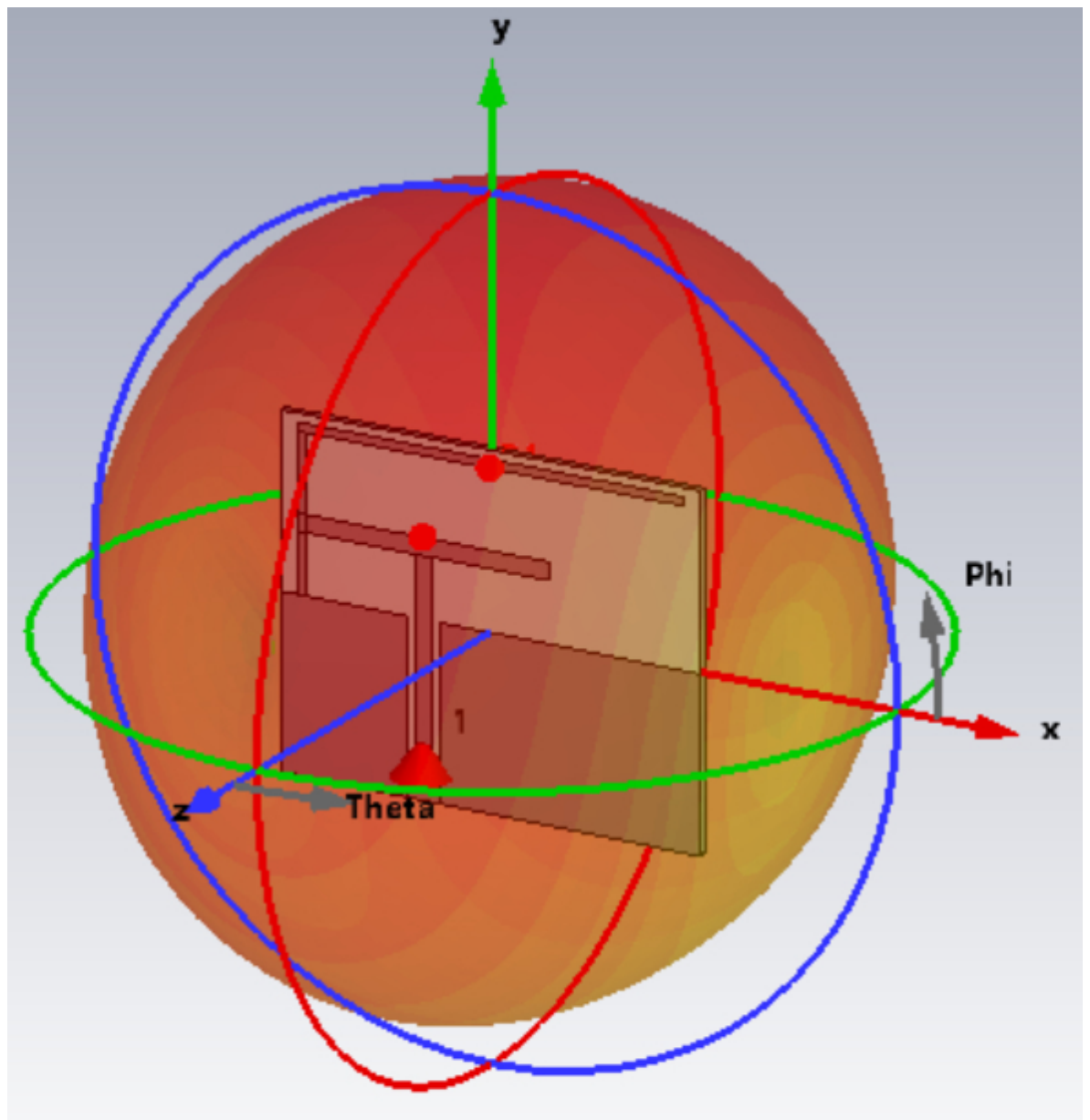}
        \caption{Dual-band IFA at 3.2 GHz}
        \label{fig:sub2}
    \end{subfigure}
    \hfill
    \begin{subfigure}[b]{0.3\textwidth}
        \centering
        \includegraphics[width=\textwidth]{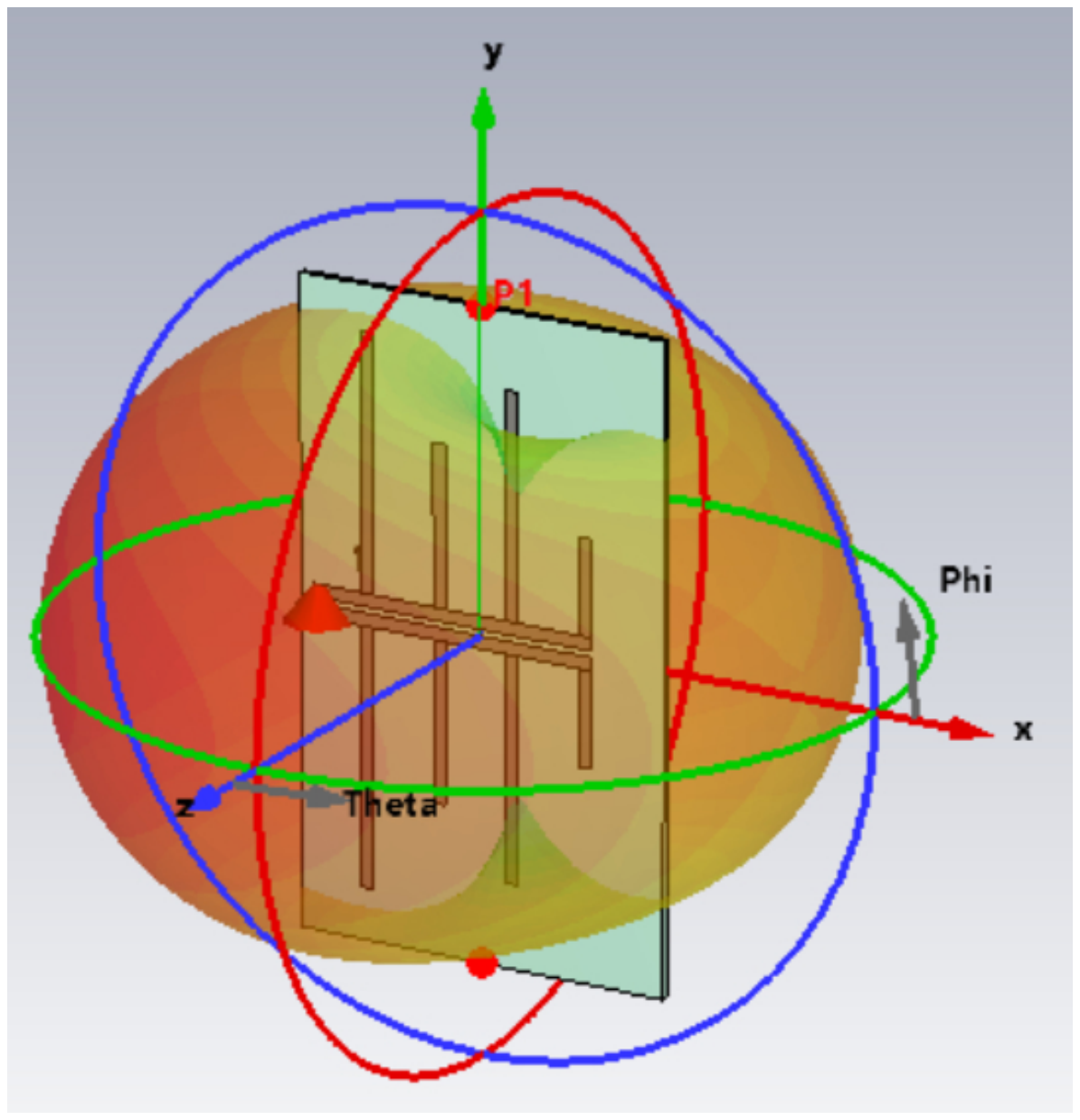}
        \caption{Multi-band dipole at 3.8 GHz.}
        \label{fig:sub3}
    \end{subfigure}
    \caption{{Radiation patterns of the antennas.}}
    \label{fig:rad_patterns}
\end{figure}

{Fig.~\ref{fig:rad_patterns} illustrates the radiation patterns of the antennas examined in this study at various frequencies. The gains of antennas in these simulations are obtained as 2.84 dBi for IFA, 3.27 dBi for dual-band IFA, and 4.77 dBi for the multi-band dipole antenna. At these operating frequencies, the total efficiencies of the antennas are obtained as 96.7\%, 91.5\%, and 98.7\%, respectively, for IFA, dual-band IFA, and multi-band dipole antennas. Note that these parameters and operating frequencies also vary as the antenna shape parameters change, since they are highly dependent on the shapes and sizes of the antenna parts.} 

These simulation results proved that gprMax can be employed for simulating more complex antenna structures in addition to traditional antennas, as shown in previous studies \cite{ref9,ref2,ref7}. Determining a finer FDTD cell size in the simulations ensures that the results match well with the ones obtained via commercial EM simulation software. However, this also substantially raises the number of iterations that need to be calculated during the simulations. Subsequently, in the next section, the performance of the gprMax with CPU and GPUs will be investigated and compared.

\begin{table*}[h]
\centering
\caption{Performance of gprMax with different FDTD cell sizes on RTX 3070 GPU}
\begin{tabular}{||c|c|c|c|c||}
\hline
\begin{tabular}[c]{@{}c@{}} FDTD Discretization\\ {[}mm{]}\end{tabular} & \begin{tabular}[c]{@{}c@{}}FDTD Model Size\\  {[}Million cells{]}\end{tabular} & \begin{tabular}[c]{@{}c@{}}GPU Ram Usage\\ {[}MB{]}\end{tabular} & \begin{tabular}[c]{@{}c@{}}Performance\\ {[}Iterations/s{]}\end{tabular} & \begin{tabular}[c]{@{}c@{}}Time\\ {[}s{]}\end{tabular} \\
\hline
\hline
dx=0.2 dy=1 dz=0.2                                                        & 7.84                                                           & 639                                                              & 376                                                                      & 170                                                    \\
\hline
dx=0.5 dy=1 dz=0.5                                                        & 1.25                                                           & 253                                                              & 1720                                                                     & 15.7                                                   \\
\hline
dx=1 dy=1 dz=1                                                            & 0.3136                                                          & 191                                                              & 3220                                                                     & 4                                                      \\
\hline
dx=0.5 dy=0.5 dz=0.5                                                      & 2.5                                                            & 321                                                              & 928                                                                      & 33.6                                                   \\
\hline
dx=0.2 dy=0.5 dz=0.2                                                      & 15.6                                                           & 1040                                                             & 195                                                                      & 338                                                    \\
\hline
dx=0.2 dy=0.2 dz=0.2                                                      & 39.2                                                           & 2260                                                             & 76.3                                                                     & 1020
\\
\hline
\end{tabular}
\label{table:PerformanceFDTD}
\end{table*}

\subsection{Hardware and Performance}

This section presents the comparison of the simulations in terms of computational performance on a high-end desktop CPU (i.e., AMD Ryzen 3900X), an entry-level professional computer-aided design (CAD) GPU (i.e., Nvidia Quadro P620), and a high-end gaming GPU (i.e., Nvidia GeForce RTX 3070). The architecture details of CPU and GPUs considered for the computational performance comparison are given in Table~\ref{table:hardware}, where it can be seen that the CPU has a very powerful but limited number of multi-purpose cores, i.e. 12 cores, while GPUs have a large number of smaller and specialized graphic processing units, namely CUDA cores, operating at lower clock rates. Interestingly, Ryzen 9 3900X and Quadro P620 have a similar number of transistors. However, they have completely different architectures, while the number of transistors utilized in RTX 3070 is more than 5 times that of the others. Moreover, this table shows that RTX 3070 has 5888 CUDA cores while Quadro P620 has 512 CUDA cores operating at a lower clock frequency compared to RTX 3070. GPU computing is mainly performed by CUDA cores. Accordingly, around 12 times or more performance difference is expected between these two GPUs. It should be noted that the GPU RAM (random-access memory) size is important for GPU computing since the entire FDTD model must be cached in GPU RAM for maximum performance and to avoid bottlenecks between the GPU and the system RAM during the simulations. Therefore, the size of the FDTD model, which can be efficiently simulated by Quadro P620, is limited to 2 GB, while RTX 3070 can simulate FDTD models with up to 8 GB. The antenna simulation framework was developed based on gprMax libraries in Python and is run on Ubuntu Mate 20.



\begin{figure}
\centering{}\includegraphics[width=0.6\columnwidth]{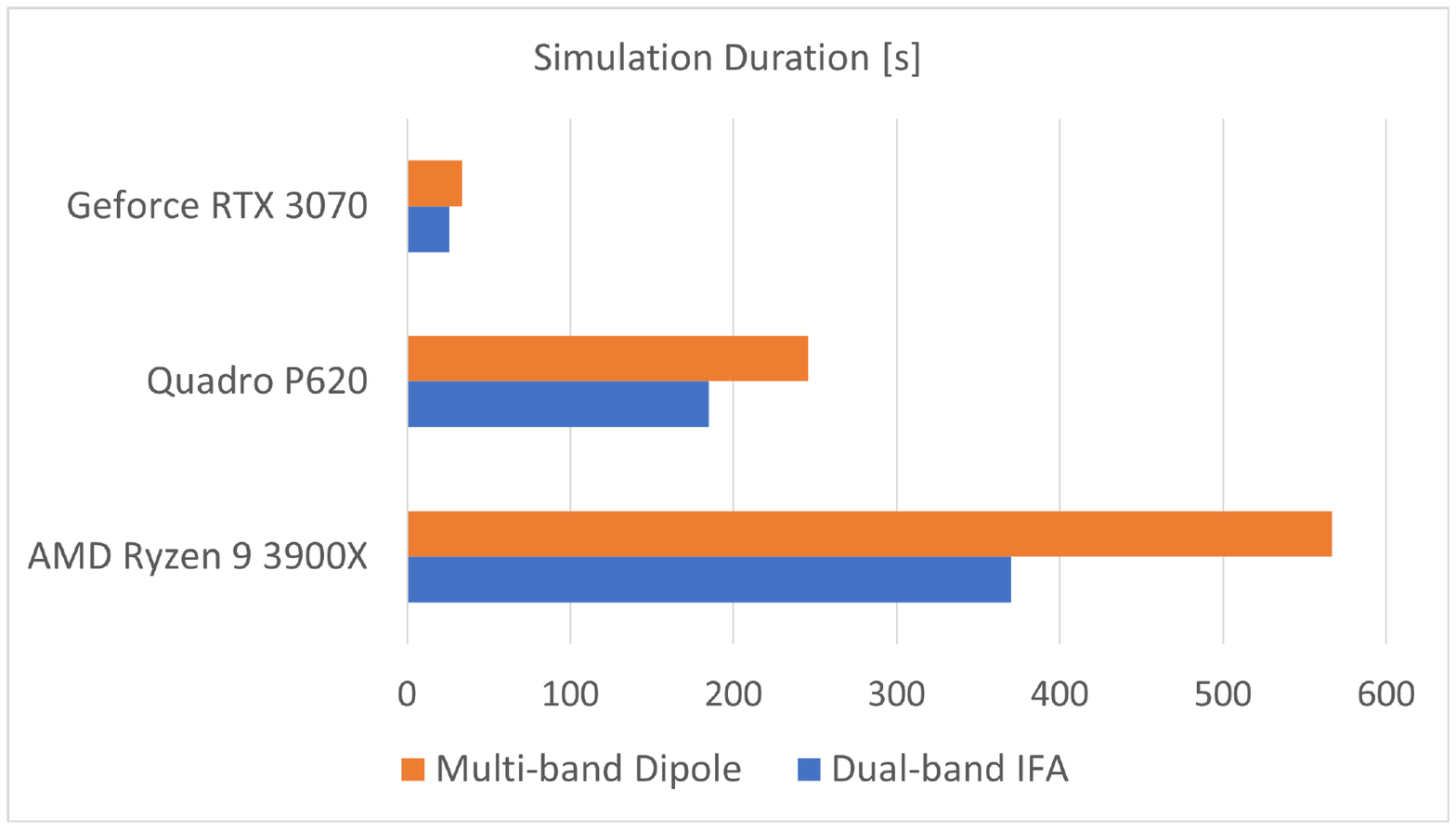}\vspace{-0.05in}\caption{Required time to complete the same simulations on AMD Ryzen 9 CPU (24 Cores with HT), Geforce Quadro P620 GPU, and Geforce RTX 3070 GPU. \label{fig:Sim_times}}
\end{figure}

\begin{figure}
\centering{}\includegraphics[width=0.6\columnwidth]{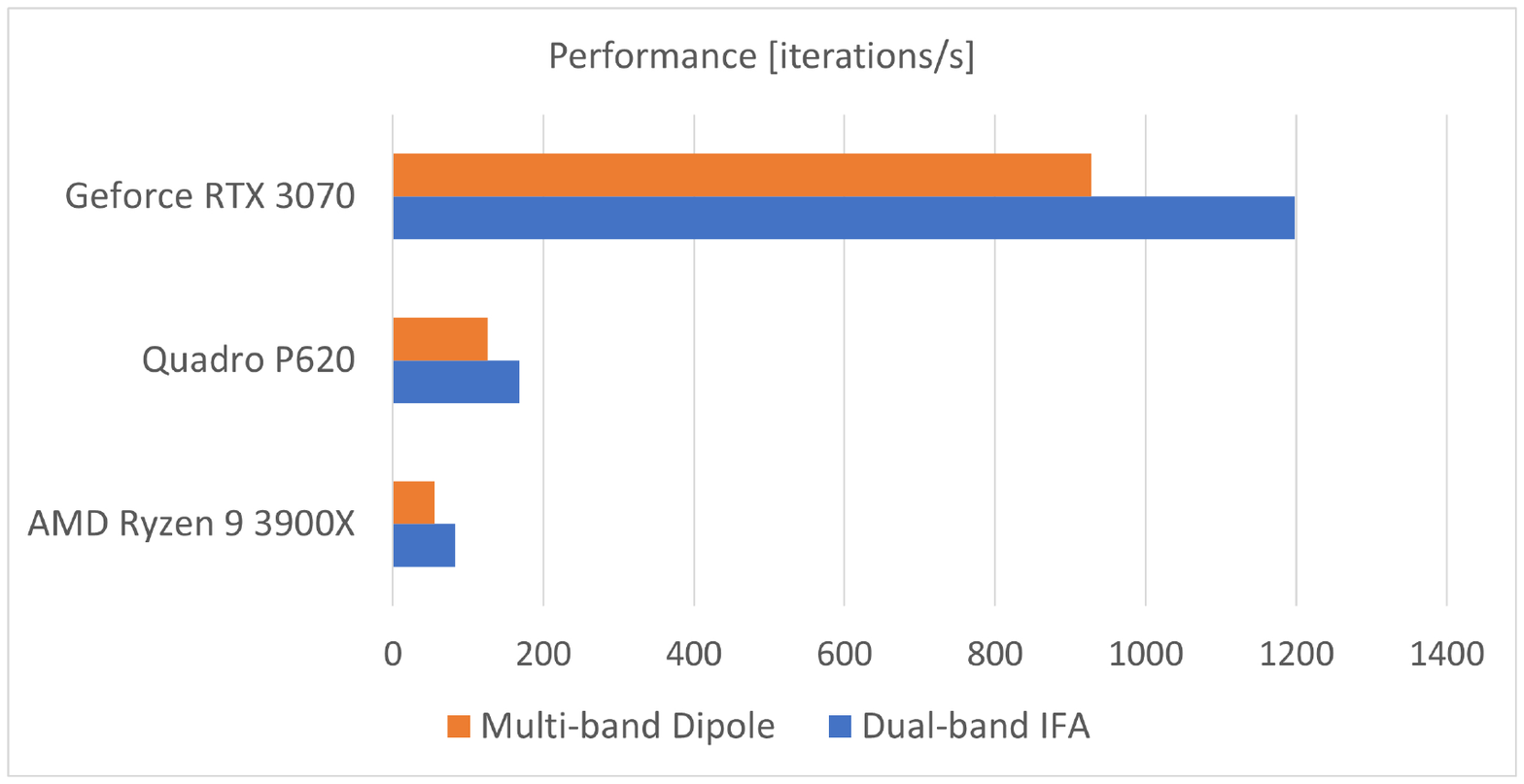}\vspace{-0.05in}\caption{Computational performance of AMD Ryzen 9 CPU (24 Cores with HT), Geforce Quadro P620 GPU, and Geforce RTX 3070 GPU. The computational performance is presented in terms of calculated FDTD iterations per second by the specified hardware. \label{fig:Sim_performance}}
\end{figure}

Open-source EM library, gprMax, uses a fixed FDTD cell size for creating the entire FDTD mesh of the simulated structures. Thus, its simulation time is mainly related to the FDTD cell size and the size of the simulated area. Accordingly, the simulation times of the single-band IFA and dual-band IFA are very similar since they have the same PCB size and the entire FDTD model size is $80\times80\times40$ mm$^3$ for both IFA models, while the model size of the multi-band dipole antenna is $80\times140\times28$ mm$^3$. Therefore, only dual-band IFA and multi-band IFA will be compared in terms of simulation performance in this section.

Fig.~\ref{fig:Sim_times} presents the simulation times of dual-band IFA and multi-band dipole antenna on the aforementioned CPU and GPUs. It shows that the multi-band dipole antenna simulation takes longer than the IFA simulation since the multi-band dipole FDTD model is larger than the IFA model. In terms of hardware comparison, RTX 3070 is substantially faster than others, and it can complete the simulations in less than 25 s, while the same simulation takes around 600 s on the CPU. The performance of Quadro P620 is between AMD Ryzen and RTX 3070. This trend is also seen in the comparison of computational performance in Fig.~\ref{fig:Sim_performance}, which shows that RTX 3070 can compute nearly 18 times faster than AMD Ryzen 9. This proves that GPU computing is much faster and more cost-effective for FDTD simulations.

The relation between the FDTD cell size, RAM usage, and simulation time is presented in  Table~\ref{table:PerformanceFDTD}, which presents the FDTD model sizes for the multi-band dipole antenna with various FDTD cell (discretization) sizes. Moreover, it also shows computation time, computation performance (iterations per second), and GPU RAM usage of these models on RTX 3070 for comparison. As previously discussed, gprMax simulations with a finer FDTD cell size (i.e., $dx=0.2$ mm, $dy=0.2$ mm, $dz=0.2$ mm) provide the most accurate results. However, this requires 2260 MB GPU RAM and the simulation duration is 1020 s (17.66 minutes) while the same model is simulated in only 4 s with a coarse FDTD cell size (i.e., $dx=1$ mm, $dy=1$ mm, $dz=1$ mm). Considering the S11 results presented in Fig.~\ref{fig:s11_ifa}, Fig.~\ref{fig:s11_dualifa} and Fig.~\ref{fig:s11_multidipole} and simulation performance listed in Table~\ref{table:PerformanceFDTD}, FDTD simulations with (i.e., $dx=0.2$ mm, $dy=0.5$ mm, $dz=0.2$ mm) provide a good trade-off between the accuracy and simulation duration up to 6 GHz frequencies such that antennas can be simulated in 338 s (5.6 minutes) with an acceptable accuracy. {As a summary, the computational complexity of FDTD simulations is heavily influenced by the cell size used for spatial discretization. As the FDTD cell size decreases, the number of grid cells increases cubically, leading to significantly higher memory usage and longer simulation times. For instance, using a fine grid of 0.2 mm in all directions results in high accuracy but requires 2260 MB of GPU RAM and a simulation time of 1020 seconds. In contrast, a coarse grid of 1 mm reduces the simulation time to just 4 seconds with minimal memory demands, but at the cost of reduced accuracy.}

For machine learning applications or antenna surrogate models, we may need to simulate 1000 or more antennas with different antenna shape parameters. Accordingly, utilizing a powerful GPU can reduce the data generation time from a few months to a few days compared to simulations performed on a CPU. The proposed framework has been used to accurately simulate 1000 multi-band dipole antennas with random shape parameters (i.e., $a1$, $a2$, $a3$ and $a4$ randomly selected within a predefined range for each parameter) with the FDTD cell size of ($dx=0.2$ mm, $dy=0.5$ mm, $dz=0.2$ mm). This simulation took around 94 hours on an RTX 3070 GPU to generate the dataset, which can be used for machine learning applications or to establish surrogate models. {The next section evaluates the machine learning methods and their prediction accuracy for antenna shape parameters using the dataset generated via the proposed framework.}

\section{{Machine Learning-based Antenna Shape Parameter Prediction}}

This section presents the machine learning-based antenna parameter prediction that utilizes the antenna data set generated using the GPU-powered simulation framework developed based on gprMax. A dual-band IFA, given in Figure~\ref{fig:dual_ifa}, was simulated in the proposed GPU-powered simulation framework with random parameters as an example application. The generated data set comprises the design parameters ($a_1$, $a_2$, $s_1$, $s_2$) and corresponding s-parameters between 0 and 6 GHz of 1800 dual-band IFAs for training machine learning models described below.

The data set is divided into two sets; the training data set comprises the shape and s-parameters of 1600 antennas, while the parameters of the remaining 200 antennas are employed to test the trained machine learning models. Various machine learning methods are used to predict the design parameters for desired operational frequency bands. The input of the machine learning models is a vector consisting of $S_{11}$ parameters between 0 and 6 GHz, while the output is the vector of design parameters ($\mathbf{w} = [a_1, a_2, s_1, s_2]$). The vector of $S_{11}$ parameters comprises 201 samples corresponding to a 30 MHz sampling resolution. {Multiple machine learning models are considered in this study to predict the antenna design parameters for the desired $S_{11}$ response. Gradient boosting frameworks and deep learning are widely used machine learning methods due to their accuracy and performance in sophisticated prediction tasks. Thus, we have chosen these methods for the prediction of antenna shape parameters, along with simpler machine learning methods, such as regression and support vector machines, to compare their accuracies. Consequently, the following machine learning methods are considered in this study.}

\subsection{{
Regression}}

The regression predicts target values with regard to input features. The regression aims to model the relationship between a dependent (target) variable and independent (input) variables  \cite{ref32}. In this study, linear regression (LR), Least absolute shrinkage and selection operator (Lasso) regression, ridge (Ridge) regression \cite{ref33}, Gaussian process regression (GPR) \cite{ref34}, and voting regression are considered to predict antenna shape parameters. 

\textbf{Linear regression:} LR assumes that there exists a linear relationship between the inputs and the output variables. Thus, it produces a linear function plus noise to model the relation between the input and output.

\textbf{ Least absolute shrinkage and selection operator:} Lasso regression is a linear regression model that improves prediction and interpretability by adding a penalty on the absolute values of the coefficients. This constraint shrinks some coefficients to zero, effectively removing less important features and performing automatic feature selection. \cite{ref35}. 

\textbf{Ridge regression:} Ridge introduces a penalty on the size of the coefficients in the model to prevent the estimated values from becoming too large, as \cite{ref36},
\begin{equation}
\min_{\mathbf{w}} \left\| \mathbf{X}\mathbf{w} - \mathbf{y} \right\|_2^2 + \alpha \left\| \mathbf{w} \right\|_2^2,
\end{equation}
where $\mathbf{X}$ denotes the matrix of input features, $\mathbf{w}$ denotes the vector of model weights (coefficients) to be learned, $\mathbf{y}$ denotes the vector of actual target values, $\alpha$ denotes the regularization parameter that controls how much penalty is added for large coefficients. This improves the accuracy, stability, and reliability of the model, especially when some input variables are correlated.

\textbf{Gaussian Process Regression:} GPR is a  Bayesian approach to model the data by defining a distribution over possible functions, by capturing patterns in the data. It is especially useful for problems with limited data  \cite{ref37}. Unlike traditional regression methods that assume a fixed form for the relation between the data, GPR defines a distribution and updates this distribution based on observed data.

\textbf{Voting Regression:} VR is an ensemble method that merges the predictions from various individual regression models to produce an improved prediction. VR averages the predictions of its constituent models, leading to more accurate and robust predictions compared to individual models \cite{ref38}. In this study,
VR incorporates gradient boosting regression, random forest regression, and linear regression methods.

\subsection{{Support Vector Machines}}

Support Vector Machines (SVMs) are powerful supervised learning models used for classification and regression. The fundamental idea behind SVMs is to find a hyperplane that separates data points of different classes with the maximum margin \cite{ref39}. For a linearly separable dataset, the optimal hyperplane can be defined as $\mathbf{w} \cdot \mathbf{x} + b = 0$, where $\mathbf{w}$ is the weight vector orthogonal to the hyperplane, $\mathbf{x}$ is the input feature vector, and $b$ is the bias term. The decision boundary is chosen such that it maximizes the margin between the two classes, and only the data points that lie closest to this boundary, which are known as support vectors, influence its position. In cases where the data is not linearly separable, SVMs employ a kernel method to map input data into a higher-dimensional space where a linear separator may exist. This is achieved using a kernel function,
\begin{equation}
   K(\mathbf{x}_i, \mathbf{x}_j) = \phi(\mathbf{x}_i) \cdot \phi(\mathbf{x}_j), 
\end{equation}
where $\phi$ is a nonlinear mapping function. The decision function in this higher-dimensional space becomes:
\begin{equation}
f(\mathbf{x}) = \sum_{i=1}^n \alpha_i y_i K(\mathbf{x}_i, \mathbf{x}) + b,  
\end{equation}
where $\alpha_i$ are Lagrange multipliers determined during training. Common kernels include the linear kernel, polynomial kernel, and the radial basis function (RBF) kernel. This method allows SVMs to perform well even with complex and nonlinearly separable data.

\subsection{Gradient Boosting}

Gradient boosting algorithms combine weak learners into a strong learner in an iterative way. The goal of gradient boosting is to find an approximation of the function $\mathbf{y}=F(\mathbf{x})$, which is a weighted sum of the weak learners, i.e., decision trees. Function $F(\mathbf{x})$ maps instances, $\mathbf{x}$, to their output values, $\mathbf{y}$ \cite{ref40}. 

In this study, we also investigated the accuracy and computational performance of three widely-used gradient boosting frameworks, namely, eXtreme Gradient Boosting (XGBoost) \cite{ref41}, Light Gradient Boosting Machine (LightGBM) \cite{ref42}, and Categorical Boosting (CatBoost) \cite{ref43} frameworks. These frameworks are developed based on gradient boosting and decision trees to provide a good generalization capability with reduced training time.

\subsection{Deep Learning}

In addition to the machine learning models given above, a deep learning model is also considered in this study. It consists of three hidden layers, each comprising 256 neurons with a ReLU (rectified linear unit) activation function. Moreover, the input and output layers of the deep learning model have 201 and 4 nodes, respectively, corresponding to the sizes of the input and output vectors. The loss function employed is the mean squared error of the predicted values given by

\begin{equation}
    \mathcal{L} = \frac{1}{N}\sum_{n=1}^{N}\left(\hat{w}_n - w_n \right)^2,
\end{equation}
where $w_n$ and $\hat{w}_n$ denote the $n$th element of the ground-truth design parameters vector and predicted design parameters vector, respectively. Moreover, the Adam optimizer is employed during training, and the TensorFlow library is used for the deep learning implementation in Python \cite{ref44}. 

\begin{figure}
\centering{}\includegraphics[width=0.7\columnwidth]{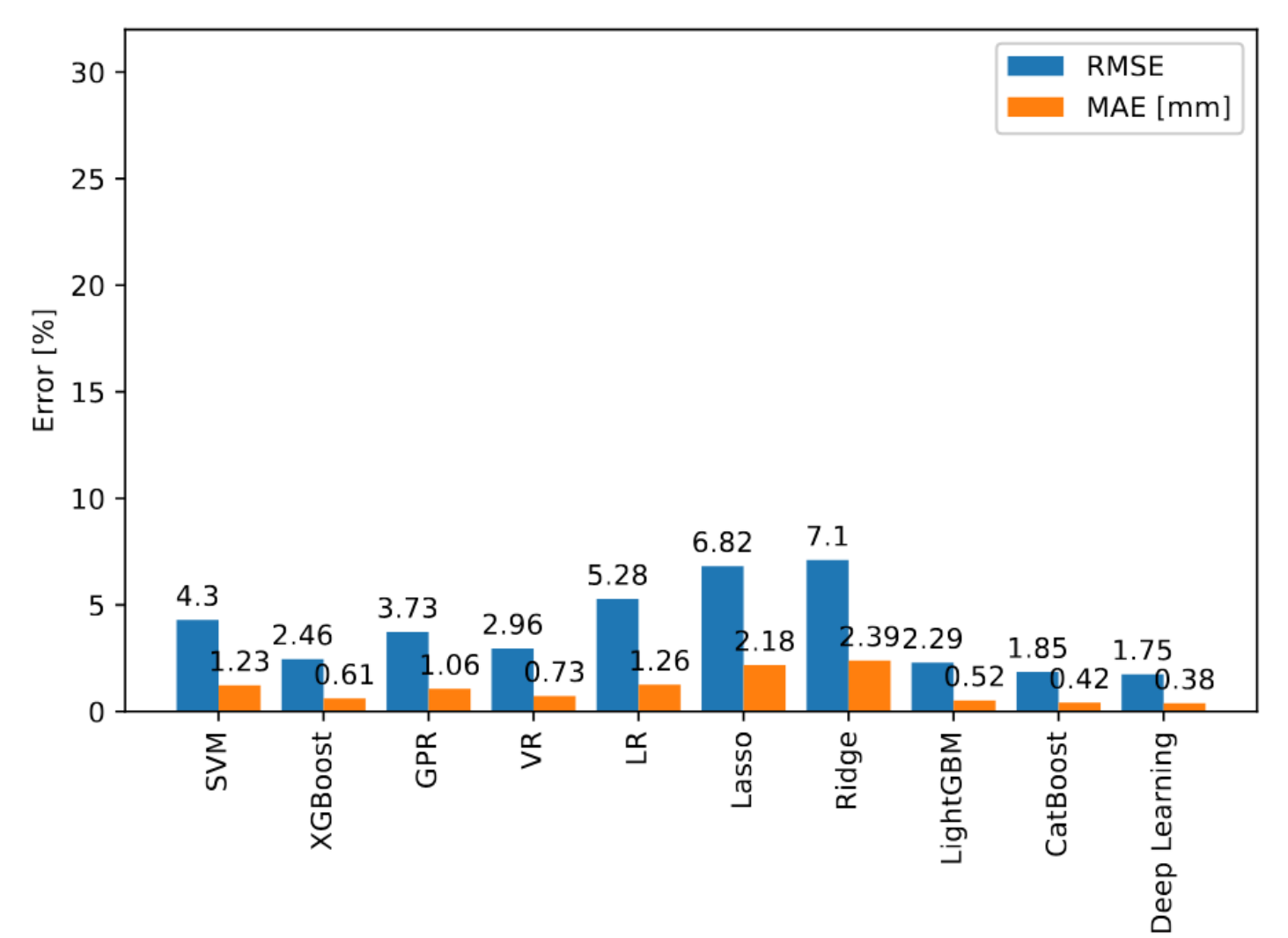}\vspace{-0.05in}\caption{Comparisons of errors (RMSE and MAE) of predicted antenna shape parameters using various machine learning techniques for a dual-band IFA antenna that is simulated using the proposed framework.  \label{fig:ML_test}}
\end{figure}

\subsection{Implementation of Machine Learning Models}

The ref45 library is utilized in Python to implement SVM, GPR, VR LR, Lasso, and Ridge. methods \cite{ref45}. The TensorFlow library is used for the deep learning implementation in Python \cite{ref44}. Although some of the machine learning models mentioned above support GPU-powered training, all machine learning methods are trained and run on an Intel i9-13900HX CPU to provide a fair comparison in terms of the training times of models. LightGBM, CatBoost, and XGBoost libraries are used to implement gradient boosting frameworks.  Although some of the machine learning models mentioned above support GPU-powered training, all machine learning methods are trained and run on an Intel i9-13900HX CPU. { Table~\ref{tab:model_params} presents the hyperparameters of the machine learning methods employed in this study.}

\begin{table}[h!]

\centering
\caption{Summary of hyperparameters of the machine learning methods employed.}
\begin{tabular}{|l|l|p{4cm}|l|}
\hline
\textbf{Model} & \textbf{Parameter} & \textbf{Explanation} & \textbf{Value} \\
\hline
\textbf{SVM} & kernel & Kernel type & poly \\
\cline{2-4}
 & epsilon & Tolerance for stopping criterion & 1e-7 \\
\hline
\textbf{Lasso} & alpha & Regularization strength & 0.001 \\
\hline
\textbf{Ridge} & alpha & Regularization strength & 0.001 \\
\hline
\textbf{Light GBM} & num\_leaves & Max leaves per tree & 100 \\
\cline{2-4}
 & learning\_rate & Step size shrinkage & 0.05 \\
\hline
\textbf{CatBoost} & iterations & Boosting iterations & 400 \\
\cline{2-4}
 & loss\_function & Loss function & RMSE \\
\hline
\textbf{DNN} & layers & Hidden layers and neurons & 3 layers, 256 neurons each \\
\cline{2-4}
 & activation & Hidden layer activation & relu \\
\cline{2-4}
 & output\_dim & Output size & 4 \\
\cline{2-4}
 & optimizer & Optimizer used & Adam \\
\cline{2-4}
 & epochs & Number of epochs & 1500 \\
\cline{2-4}
 & batch\_size & Batch size & 100 \\
\hline
\end{tabular}
\label{tab:model_params}
\end{table}

\subsection{Accuracy of Machine Learning Models}

As a performance metric to measure the accuracy of predictions, root mean squared error (RMSE) is used as given by,
\begin{equation}
\text{RMSE} = \sqrt{\frac{1}{N} \sum_{n=1}^{N} \left( \hat{y}_n - y_n \right)^2},    
\end{equation}
where \( N \) denotes the number of data points, i.e., the total number of predictions, \( \hat{y}_n \) denotes the predicted value of the \( n \)-th prediction and \( y_n \) denotes the actual (true) value of the \( n \)-th prediction of the antenna shape parameters. Another performance metric is the mean absolute error (MAE) of the predicted shape parameter in comparison with the ground truth values in millimeters.

Figure~\ref{fig:ML_test}
shows the antenna shape parameter prediction accuracy of the machine learning models are given above. It shows that deep learning outperforms other methods and achieves the minimum prediction error overall, while CatBoost provides the best prediction accuracy after deep learning. Other gradient boosting methods follow these methods in terms of estimation accuracy. Since VR also utilizes random forest, it also performs better than regression methods and SVM. The relation between the antenna shape parameters and S-parameters is sophisticated and highly nonlinear; hence, SVM and regression methods cannot learn this relationship sufficiently to provide accurate predictions. These results show deep learning is the most suitable method to estimate antenna shape parameters, while the second-best method is gradient boosting frameworks, especially CatBoost.

\section{Conclusion}\label{conclusion}
This study has presented a method for the simulation of antennas via gprMax on GPUs and compared the performance of CPU and GPU computing for EM simulations of three different antenna models. Furthermore, it has been shown that gprMax can be effectively used in the simulation of complicated antenna structures that may have multiple resonance frequencies, such as dual-band IFA or multiband dipole antennas. The simulation results demonstrated that gprMax with a fine FDTD cell size can achieve a satisfactory accuracy level that is comparable to the accuracy of commercial EM simulation software packages. It is evident that utilizing GPUs in EM simulations can substantially accelerate the FDTD simulations. Therefore, GPUs are especially required when simulating large antenna structures with a fine FDTD cell size or generating large antenna data sets for machine learning or surrogate modeling applications. In future work, we will study machine learning and surrogate models for antenna design and optimization using the data sets generated via the method described in this study.

\subsection{Acknowledgements} This study was supported by the European Innovation Council under the Innovation in SMEs (INNOSUP) program and in part by METU
BAP Grant AGEP-301-2025-11558.



\end{document}